%% file: main.tex
\documentclass[conference]{IEEEtran}
\usepackage{times}
\usepackage{graphicx}
\usepackage{tikz}
\usetikzlibrary{arrows.meta,positioning,fit,calc}
\usepackage{amsmath}
\usepackage{amssymb}
\usepackage{booktabs}
\usepackage{subcaption}
\usepackage{cuted}
\usepackage{todonotes}

\usepackage[numbers]{natbib}
\usepackage{multicol}
\usepackage[bookmarks=true]{hyperref}

\begin{document}

\title{Sound of Touch: Active Acoustic Tactile Sensing via String Vibrations}


\author{
\IEEEauthorblockN{
Xili Yi\IEEEauthorrefmark{1},
Ying Xing\IEEEauthorrefmark{1},
Zachary Manchester\IEEEauthorrefmark{2},
Nima Fazeli\IEEEauthorrefmark{1}
}
\IEEEauthorblockA{
\IEEEauthorrefmark{1}University of Michigan\\
\IEEEauthorrefmark{2}MIT
}
}


%


\maketitle

\begin{strip}
    \centering
    \includegraphics[width=0.99\linewidth]{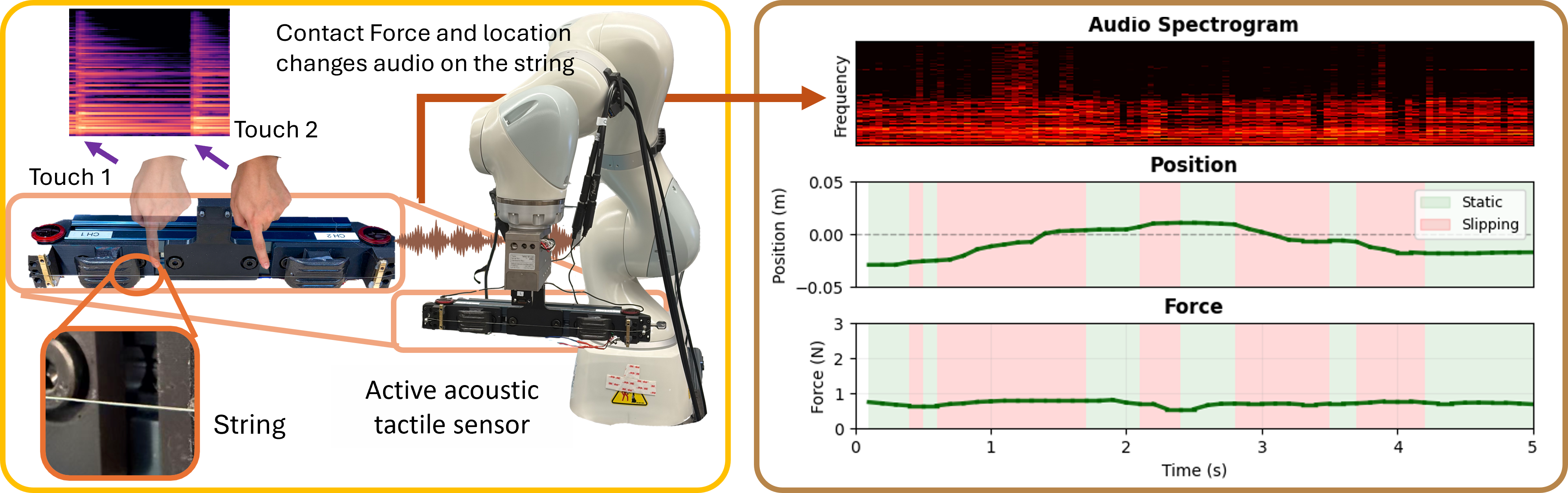}
    \captionof{figure}{Sound of Touch leverages vibration as tactile sensing. Contact with a continuously excited string produces characteristic spectral changes that encode contact location, force, and slip, enabling real-time tactile inference from short-duration audio signals.} 
    \label{fig:teaser}
\end{strip}

\begin{abstract}
Distributed tactile sensing remains difficult to scale over large areas: dense sensor arrays increase wiring, cost, and fragility, while many alternatives provide limited coverage or miss fast interaction dynamics. We present \textit{Sound of Touch}, an active acoustic tactile-sensing methodology that uses vibrating tensioned strings as sensing elements. The string is continuously excited electromagnetically, and a small number of pickups (contact microphones) observe spectral changes induced by contact.
From short-duration audio signals, our system estimates contact location and normal force, and detects slip. To guide design and interpret the sensing mechanism, we derive a physics-based string-vibration simulator that predicts how contact position and force shift vibration modes. Experiments demonstrate millimeter-scale localization, reliable force estimation, and real-time slip detection.
Our contributions are: (i) a lightweight, scalable string-based tactile sensing hardware concept for instrumenting extended robot surfaces; (ii) a physics-grounded simulation and analysis tool for contact-induced spectral shifts; and (iii) a real-time inference pipeline that maps vibration measurements to contact state.
\end{abstract}

\IEEEpeerreviewmaketitle

\input{text/1_introduction}

\input{text/2_relatedwork}

\input{text/3_methodology}

\input{text/4_system_design}

\input{text/5_experiment}

\input{text/6_results}

\input{text/7_limitation}

\section*{Acknowledgments}


\bibliographystyle{plainnat}
\bibliography{references}

\clearpage

\input{Appendix}

\end{document}

%% file: text/1_introduction.tex
\section{Introduction}

Robust physical interaction requires tactile feedback. In uncertain environments, touch lets robots detect contact location, transmitted forces, and relative motion (e.g., incipient slip), enabling timely reactions to intended or unintended contacts. Yet today’s tactile sensing remains a bottleneck. High-performance sensors are typically confined to small fingertip patches, while extending sensing beyond the fingertip to robot links and tools is limited by wiring complexity, cost, durability, and form-factor constraints. This leaves a critical gap: a lightweight, scalable tactile modality that can provide distributed contact information over extended structures while still capturing the fast, high-frequency signatures of dynamic events such as slip.

To overcome these limitations, recent work has explored alternative tactile sensing strategies: Vision-based tactile sensors 
such as \textit{GelSight~}\cite{yuan2017gelsight}, \textit{Gelslim}~\cite{donlon2018gelslim, taylor2021gelslim3, sipos2024gelslim}, \textit{Soft Bubbles}~\cite{alspach2019soft, kuppuswamy2020soft}, \textit{Omnitact}~\cite{padmanabha2020omnitact}, and  \textit{FingerVision}~\cite{yamaguchi2017implementing, yamaguchi2019recent} provide detailed contact geometry and deformation fields but require cameras, illumination, and transparent elastomers that constrain form factors and integration. Flexible resistive, capacitive, or magnetic tactile arrays can cover larger areas, such as \textit{Xela} uskin~\cite{tomo2016modular, tomo2017covering}, \textit{AnySkin}~\cite{bhirangi2025anyskin} and \textit{eFlesh}~\cite{pattabiraman2025eflesh}, yet their spatial resolution, shear measurement, signal stability, and lifetime often pose challenges.
Vibration- and acoustic-based sensing has also been used to detect contact onset, slip, and material properties by exploiting high-frequency signals that encode rich interaction dynamics~\cite{jamali2010material, fernandez2014micro, suhn2023vibro, lu2023active}. However, prior approaches assume rigid sensor structures and do not provide spatially distributed tactile information along extended bodies.

In this work, we explore a tactile sensing modality based on the vibration response of tensioned strings, inspired by string instruments such as guitar, as shown in Fig. \ref{fig:teaser}. A taut string behaves as a one-dimensional
continuous sensing element whose resonant spectra vary systematically with the location and depth of contact, analogous to how touching a guitar string alters its pitch and timbre. By continuously exciting the string and recording vibration signals at a small number of pickup locations, we can infer contact location and applied normal force from spectral changes alone. Moreover, dynamic events like slip inject distinctive high-frequency vibration components into the signal, enabling direct detection of slippage without explicit force or motion sensing.

We introduce \textit{Sound of Touch}, an active acoustic string-based tactile sensor prototype and inference framework that combines physical modeling, simulation-based analysis, and real-world audio processing. We develop a detailed string-vibration simulator to provide a physics-grounded understanding of how contact position and press depth modulate vibration modes, while using supervised learning to map real audio measurements to contact quantities. Our contributions are threefold: (1) a scalable tactile sensing concept using tensioned strings and audio measurements; (2) a physics-based analysis tool for understanding vibration-mediated contact sensing; and (3) a real-time inference pipeline that estimates contact location, force, and slip from short-duration audio signals using lightweight learning models.

%% file: text/2_relatedwork.tex
\section{RELATED WORK}

\noindent\textbf{Tactile Sensing for Manipulation.}
Tactile sensing remains a critical integration bottleneck for robotic manipulation. Traditional high-resolution sensing has been dominated by vision-based tactile sensors (VBTS). Sensors such as \textit{GelSight} \cite{yuan2017gelsight}, \textit{GelSlim} \cite{donlon2018gelslim, taylor2021gelslim3}, and \textit{Soft-Bubble} \cite{kuppuswamy2020soft} utilize internal cameras to image the deformation of a soft elastomer surface, enabling millimeter-scale reconstruction of contact geometry, normal and shear forces, and incipient slip \cite{li2025classification, zhao2025universal}. \textit{FingerVision} \cite{yamaguchi2017implementing, yamaguchi2019recent} offers an alternative by using transparent skins to track markers for force estimation while providing visual proximity data. To address the cost and complexity of these systems, the \textit{DIGIT} platform \cite{lambeta2020digit} introduced a compact, mass-producible design, which has evolved into multimodal fingertips like \textit{Digit 360} \cite{lambeta2024digitizing}, integrating microphones to capture high-frequency acoustic signatures alongside visual data.

To instrument larger robot surfaces beyond the fingertips, researchers have explored flexible ``electronic skins'' using resistive, capacitive, or magnetic principles \cite{cirillo2021tactile, meribout2024tactile}. Recent advances like \textit{AnySkin} \cite{bhirangi2025anyskin} and \textit{eFlesh} \cite{pattabiraman2025eflesh} utilize magnetic fields to provide robust, zero-shot replaceable tactile modules. Despite their versatility, scaling these arrays to complex geometries often suffers from significant wiring complexity and signal drift \cite{meribout2024tactile}. Our active acoustic tactile sensor overcomes these limitations by utilizing a resonant 1D waveguide, providing distributed coverage with minimal electronic complexity.

\vspace{3pt}
\noindent\textbf{Vibration and Acoustic Sensing in Robotics.}
Vibration and acoustic modalities offer high temporal resolution, capturing dynamic events that are typically aliased by the frame rates of vision sensors \cite{suhn2023vibro, fernandez2014micro}. Prior work has used contact microphones and accelerometers to classify material textures \cite{jamali2010material} and recognize interactions through structural vibrations \cite{suhn2023vibro, taunyazov2021extended}. Active acoustic sensing extends this by injecting controlled vibrations into the system to infer contact states or object properties \cite{lu2023active, zhang2025vibecheck}. For instance, \textit{VibeCheck} \cite{zhang2025vibecheck} utilizes active signals for contact-rich manipulation, while \textit{SonicSense} \cite{liu2024sonicsense} leverages in-hand acoustic vibrations to perceive object properties during interaction.

Crucially, most purely vibro-acoustic sensors are limited in their ability to measure force. While systems like \textit{VibroTouch} \cite{suhn2023vibro} attempt to estimate light contact forces through resonant-peak shifts, they often struggle to decouple force magnitude from contact location in unstructured media. Our approach addresses this by leveraging the physics of a continuously excited tensioned string—inspired by the \textit{EBow} \cite{heet1978ebow}—to establish a continuous sensing element where normal force and contact location have distinct, decoupled spectral signatures.

\vspace{3pt}
\noindent\textbf{Slippage Detection.}
Slip detection is a canonical application for high-frequency tactile sensors \cite{fernandez2014micro, heyneman2016slip}. Early studies demonstrated that micro-vibrations generated during the transition from static to sliding friction provide reliable cues for detecting incipient slip \cite{howe1989sensing, fernandez2014micro}. While VBTS can detect slip by tracking marker displacements or surface optical flow \cite{taylor2021gelslim3, funk2024evetac}, they are frequently limited by latency. Learning-based frameworks have recently targeted universal slip detection \cite{zhao2025universal}, emphasizing the need for representations that capture high-frequency spectro-temporal dynamics \cite{suhn2023vibro}. Our sensor exploits the sensitivity of string vibrations to provide real-time, lightweight slip detection using simple regression models.

\vspace{3pt}
\noindent\textbf{Summary.}
In summary, while vision-based sensors provide high resolution, they remain difficult to scale and integrate into irregularly shaped robot surfaces. Acoustic sensing provides access to rich dynamics but has historically lacked reliable force-estimation capabilities in distributed form factors. This work addresses these gaps by leveraging continuously excited string vibrations as a physically interpretable substrate for distributed inference of contact location, force, and slippage.

%% file: text/3_methodology.tex
\section{Methodology}

\subsection{Problem Statement}

\begin{figure}
    \centering
    \includegraphics[width=0.95\linewidth]{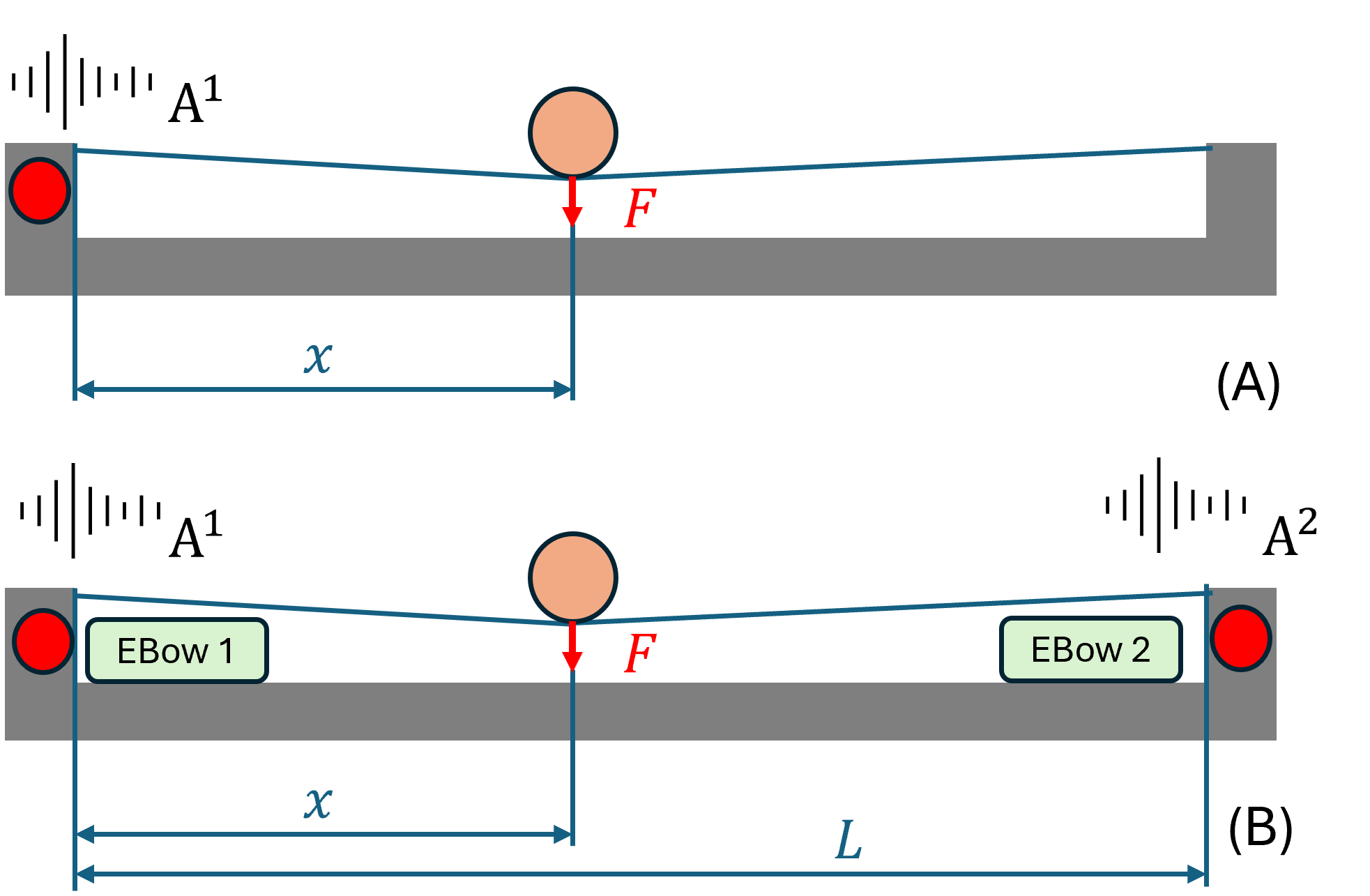}
    \caption{(A) A contact at position $x$ with force $F$ alters the string’s effective vibrating lengths and tension, inducing characteristic shifts in modal frequencies.
(B) Dual EBow-like electromagnetic drivers provide continuous excitation, while two pickup microphones capture complementary vibration responses that enable separation of contact location and force.}
    \label{fig:concept_design}
    \vspace{-4mm}
\end{figure}

We consider a string-based tactile sensor that is continuously excited by electromagnetic drivers (EBows) and instrumented with two pickup microphones, as shown schematically in Fig.~\ref{fig:concept_design}. When an object interacts with the string, the resulting vibration produces a two-channel audio signal $A = [A^1, A^2]$. The spectral structure of this signal encodes whether contact occurs, where along the string the contact is made, the applied normal force, and whether the contact is sticking or slipping. Given real-world audio measurements $A$, our goal is to infer the following contact features:
\begin{equation}
(c, x, F, s) = \mathcal{M}(A),
\end{equation}
where $c \in \{0,1\}$ denotes contact occurrence, $x$ the contact location along the string, $F$ the applied normal force, and $s \in \{0,1\}$ indicates slip versus stick. Here, the mapping $\mathcal{M}(\cdot)$ represents the desired perception model.

\subsection{Active acoustic tactile sensor design}
The core idea of our string-based tactile sensor builds on the relationship between a string’s modal frequencies, its effective vibrating length, and its tension:
\begin{equation}
f_n = \frac{n}{2L} \sqrt{\frac{T}{\mu}}, \quad n = 1, 2, 3, \ldots
\label{eq:string_harmonics}
\end{equation}
where $L$ is the effective vibrating length, $T$ is string tension, and $\mu$ is the mass density. Eqn.~\eqref{eq:string_harmonics} motivates the design shown in Fig.~\ref{fig:concept_design}(A): a taut string is fixed at both ends and when a contact occurs, a microphone measures the resulting audio signal from which contact information is inferred.

There are two important practical challenges to address: First, changes in effective length and changes in tension (caused by contact force) can shift resonant frequencies in similar directions, making it difficult to disentangle contact location from force-related effects when observing only a single measurement. Second, the system must be continuously excited to produce a stable vibration response from which spectral features are extracted.

To address both issues, we use continuous, contactless electromagnetic excitation and multi-point sensing. Specifically, we propose two EBow-like electromagnetic drivers and two pickup microphones, as shown in Fig.~\ref{fig:concept_design}(B). Dual EBows excitation is strictly necessary to ensure both string segments vibrate independently after contact partitions them. The EBow ~\cite{heet1978ebow} provides sustained excitation through electromagnetic feedback, while dual contact microphones provide complementary views of wave propagation that help disambiguate contact location from force-related spectral shifts.

\begin{figure}[]
    \centering
    \includegraphics[width=1.0\linewidth]{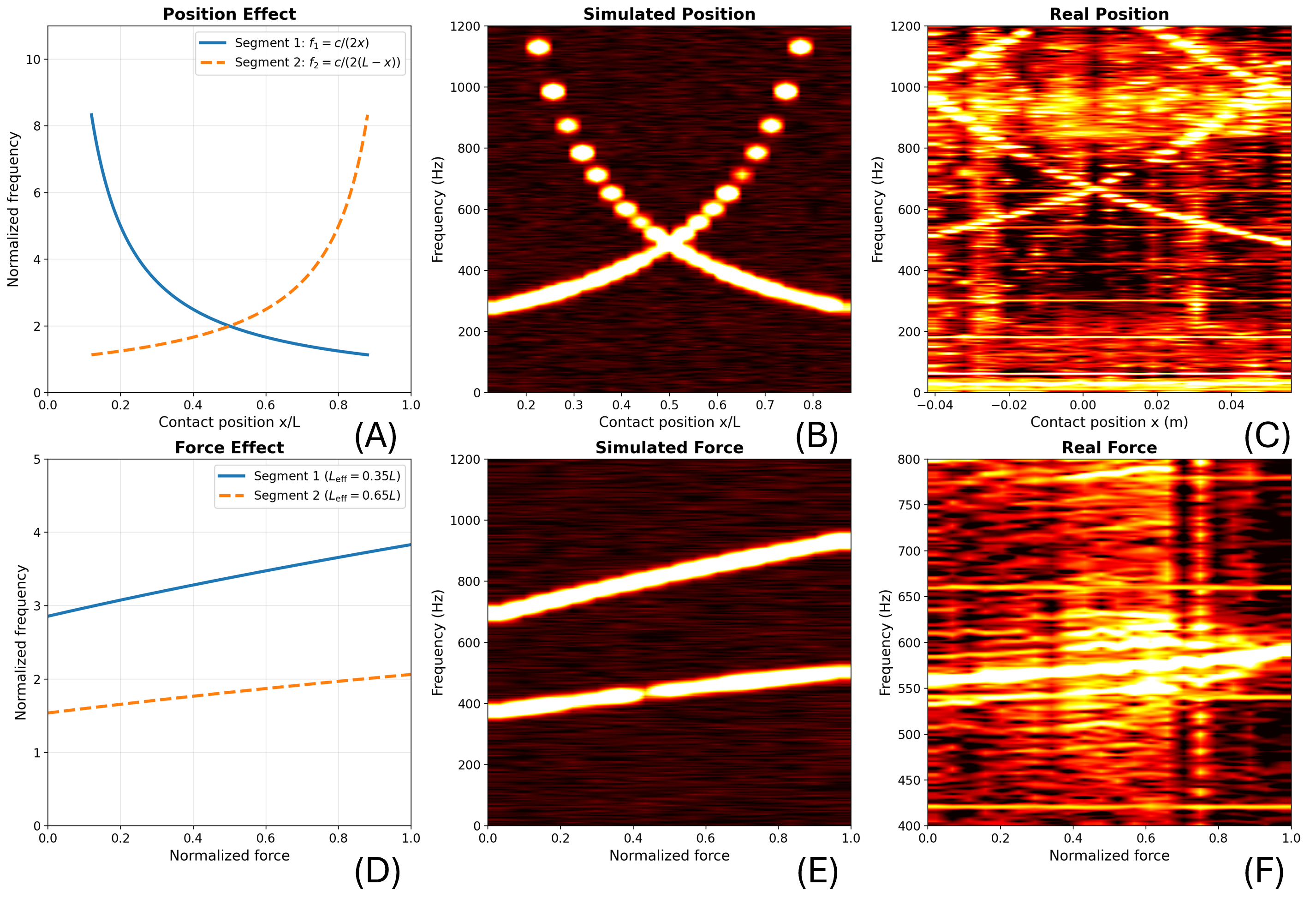}
    \caption{\textbf{Resonant-frequency modulation by contact.} (A, D) Theoretical trends showing asymmetric frequency shifts for position and monotonic shifts for force. (B, E) Simulated spectra reproducing these modal shifts. (C, F) Real-world measurements exhibiting complex harmonic structures and overtones.}
    \label{fig:frequency}
    \vspace{-6mm}
\end{figure}

Let the total string length be $L$. When a contact is applied at position $x$, the string is effectively divided into two segments of lengths $x$ and $L-x$. As the contact location moves along the string, one effective length increases while the other decreases, as illustrated in Fig.~\ref{fig:frequency}(A). Referring to Eqn.~\ref{eq:string_harmonics}, this asymmetric change in effective length leads to an increase in the modal frequencies associated with one segment and a decrease in those of the other. In contrast, increasing the pressing force raises the overall string tension, resulting in an upward shift of the frequencies on both sides, as shown in Fig.~\ref{fig:frequency}(B). This design enables a clear decomposition of the effects of contact location and applied force through their distinct and decoupled spectral signatures.

Continuous excitation is critical for extracting stable spectral features from string vibrations. In the absence of sustained driving, vibrations induced by contact decay rapidly, leading to transient responses that are highly sensitive to initial conditions and contact timing. To address this, we employ EBow-style electromagnetic drivers \cite{heet1978ebow}, which provide contactless, continuous excitation through feedback between a sensing and a driving coil. The driver induces oscillations by injecting energy at the string’s instantaneous vibration frequency, compensating for damping losses to maintain steady-state vibrations.

From a modeling perspective, the EBow can be approximated as a spatially localized forcing term $f_{\text{drive}}(x,t)$ that continuously injects energy into the system. This sustained excitation ensures that contact-induced changes in effective length and tension manifest as consistent shifts in harmonic structure, enabling reliable extraction of frequency-domain features over short time windows.

\subsection{Audio simulation}


To better connect the concept design in Sec.~III-B to the spectral trends we exploit in learning, we develop a one-dimensional string-vibration simulator. Rather than perfectly matching the full real-world audio pipeline, the simulator serves as an analysis and visualization tool to check whether the basic design intuition holds under an idealized string model. We use it to examine (i) how contact location produces asymmetric frequency shifts across the two string segments, and (ii) how increased press depth and normal force (captured through effective tension and damping) changes the observed spectra. Importantly, simulated data are not used for training or inference. Full details are provided in the Appendix.




\subsubsection{Split string approximation}
Instead of modeling the contact as a local damper interaction, we adopt a simplified but interpretable approximation: a pressing contact at location $x\in(0,L)$ is treated as a clamp,
\begin{equation}
    y(x,t)=0,
\end{equation}
which partitions the original string of length $L$ into two independent segments with effective lengths
\begin{equation}
    L_1=x, \qquad L_2=L-x.
\end{equation}
We simulate the two segment contributions as ideal source signals $A_1(t)$ and $A_2(t)$, defined as the transverse velocity measured at a chosen pickup location on each segment. In the real setup, however, each contact microphone is not perfectly isolated to one side; instead, the measured two-channel signals are modeled as a linear mixture of the two segment contributions:
\begin{equation}\label{eq:mixing}
    \begin{bmatrix}
    B_1(t)\\
    B_2(t)
    \end{bmatrix}
    =
    \begin{bmatrix}
        \lambda_1 & 1-\lambda_1\\
        1-\lambda_2 & \lambda_2
    \end{bmatrix}
    \begin{bmatrix}
        A_1(t)\\
        A_2(t)   
    \end{bmatrix}.
\end{equation}
where $\lambda_1, \lambda_2\in[0,1]$ capture pickup-specific sensitivity and cross-coupling. We use $B_1, B_2$ as the simulated microphone channels when generating spectra and features.

\subsubsection{Force-tension coupling}
To incorporate normal force, under small deformation assumption, we introduce a monotonic mapping from applied force $F$ to an effective tension:
\begin{equation}\label{eq:T_of_F}
    T(F)=T_0+k_T F,
\end{equation}
where $T_0$ is the baseline tension without pressing and $k_T$ is a tunable constant used to generate the expected trend (larger $F$ yields larger effective tension and thus higher frequencies). In addition, stronger contact typically increases energy loss. Rather than a detailed model, we fold this effect into a segment damping coefficient that can differ on the two sides,
\begin{equation}\label{eq:d_of_F}
    d_k(F)=d_{0,k}+k_{d,k}F, \qquad k \in\{1,2\}.
\end{equation}
To keep the simulator minimal while still capturing the basic physics, peak locations shift with $T(F)$, and peak magnitudes can be attenuated through $d_k(F)$.

\subsubsection{Discretized dynamics used for the audio}
For each segment $k\in\{1,2\}$, the transverse displacement $y_k(\xi,t)$ (where $\xi\in[0,L_k]$ is the local coordinate on that segment) is modeled by a damped driven wave equation
\begin{equation}\label{eq:wave_segment}
    \mu\, \frac{\partial^2 y_k}{\partial t^2}
    =
    T(F)\, \frac{\partial^2 y_k}{\partial \xi^2}
    -
    d_k(F)\, \frac{\partial y_k}{\partial t}
    +
    f_{\text{drive}}(\xi,t),
\end{equation}
with fixed boundary conditions at the segment endpoints,
\begin{equation}
    y_k(0,t)=0, \qquad y_k(L_k,t)=0.
\end{equation}
We discretize each segment into $N$ nodes with spacing $\Delta\xi=L_k/(N-1)$.
For an interior node $i$, we use the centered Laplacian and obtain
\begin{equation}
    \ddot y_i
    =
    v^2\frac{y_{i+1}-2y_i+y_{i-1}}{\Delta\xi^2}
    +\frac{F^{\mathrm{drv}}_i(t)}{\mu\Delta\xi}
    -\frac{d_k(F)}{\mu\Delta\xi}\dot y_i,
\end{equation}
where $v=\sqrt{T(F)/\mu}$ is the wave speed and $F^{\mathrm{drv}}_i(t)$ is the discrete driving force applied at node $i$.
In implementation, the drive is constructed as a spatially localized profile multiplied by a time-varying oscillation:
\begin{equation}
    F^{\mathrm{drv}}_i(t)
    =
    A_{\mathrm{drv}}\sin\!\big(2\pi f_{\mathrm{drv}}(t)\,t\big)\; g(x_i;\,x_{\mathrm{drv}},\sigma_{\mathrm{drv}}),
\end{equation}
where $g(\cdot)$ is a unit-peak localized shape centered at $x_{\mathrm{drv}}$.
To mimic the EBow feedback behavior, we update the instantaneous drive frequency $f_{\mathrm{drv}}(t)$
by estimating the dominant vibration frequency from a short sliding window of the simulated microphone signal and feeding it back to the driver:
\begin{equation}
    f_{\mathrm{drv}}(t)\leftarrow \hat f(t),
    \qquad\
    \hat f(t)=\arg\max_f \left|\mathrm{FFT}\{B_1\}_{\text{recent window}}\right|.
\end{equation}

Time integration is performed using a central-difference scheme under a constraint
\begin{equation}
    \Delta t < \alpha\,\frac{\Delta \xi}{v}
\end{equation}
with $\alpha\in(0,1)$ is chosen to ensure stability.

\subsubsection{Predicted frequency trends and feature extraction}
Using the ideal string relation in Eq.~(2), the split string model suggests two dominant feature frequencies:
\begin{equation}\label{eq:two_features}
    f^{(1)}(x,F)\approx \frac{1}{2x}\sqrt{\frac{T(F)}{\mu}},\quad
    f^{(2)}(x,F)\approx \frac{1}{2(L-x)}\sqrt{\frac{T(F)}{\mu}}.
\end{equation}
These relations yield two qualitative predictions that we use for validation: (i) at fixed $F$ (constant $T$), moving the contact location changes $x$ and $L-x$ in opposite directions, so one feature increases while the other decreases. (ii) at fixed $x$, increasing $F$ increases $T(F)$, shifting both features upward together. In simulation, we obtain $B_1(t)$ and $B_2(t)$, compute windowed FFT spectra after discarding initial transients, and track the two dominant peaks as the feature frequencies. Simulation results are shown in Fig. \ref{fig:frequency}(B), consistent with our concept design, though also exhibiting higher-order vibration modes. 


We also visualize the data distribution using t-SNE \cite{maaten2008visualizing} in Fig. \ref{fig:data}(A), a dimensionality reduction technique that embeds high-dimensional audio features into a two-dimensional space while preserving local neighbor relationships. The resulting embeddings reveal clear structural trends, suggesting that the relationship between vibration patterns and tactile states can be captured by relatively simple network architectures.



\subsection{Learning formulation}
\label{sec:learning}

\begin{figure}[h]
    \centering
    \includegraphics[width=1.0\linewidth]{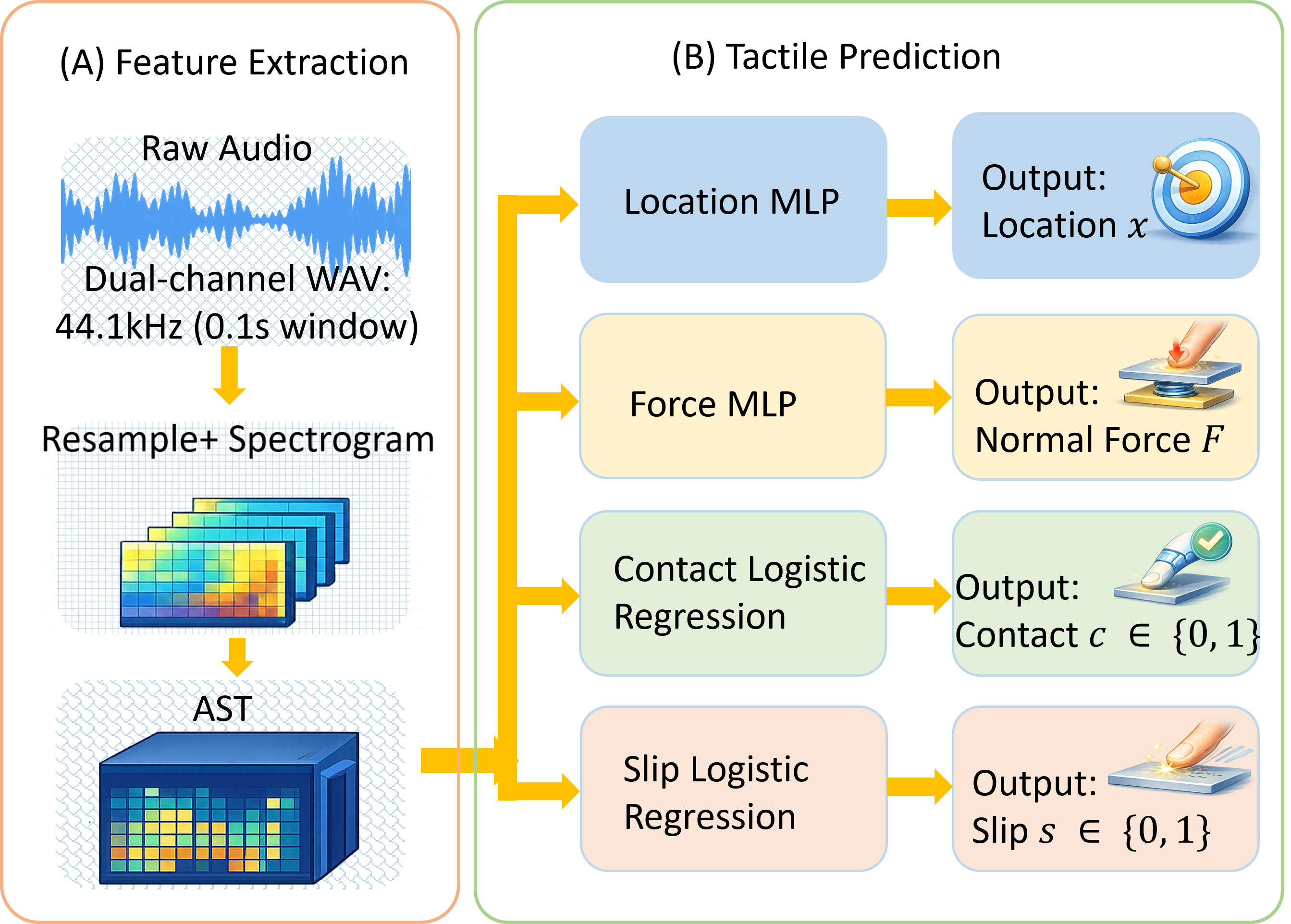}
    \caption{\textbf{Learning architecture.} Short audio windows are transformed into spectral features and encoded by a frozen Audio Spectrogram Transformer (AST). The resulting embeddings are shared by lightweight task-specific heads for contact location, normal force, contact detection, and slip detection. 
    }
    \label{fig:learning}
\end{figure}

Our learning objective is to map the raw dual-channel audio signal $A \in \mathbb{R}^{2 \times T}$ (where $T$ is the window length) to the contact state tuple $(c, s, x, F)$. We adopt a data-efficient approach that leverages the strong inductive biases of pre-trained audio models, allowing us to focus the learning budget on task-specific mappings rather than feature extraction (Fig.~\ref{fig:learning}).

\vspace{2mm}
\noindent \textbf{1) Feature Extraction Backbone:} Instead of training a feature extractor from scratch, we utilize the Audio Spectrogram Transformer (AST) pre-trained on AudioSet. The raw audio $A$ is first converted into a Mel-spectrogram representation $S \in \mathbb{R}^{128 \times 100}$. This spectral input is tokenized and passed through the AST encoder $\Phi_{AST}(\cdot)$.

We treat the AST as a frozen feature extractor. By freezing the encoder weights, we prevent overfitting to the specific acoustic properties of the lab environment and ensure that the model relies on generalized spectro-temporal features (e.g., harmonic shifts and transient bursts) that are robust across different domains. The resulting embedding $z = \Phi_{AST}(S)$ serves as a compact, high-level representation of the string's vibration state.

\vspace{2mm}
\noindent \textbf{2) Task-Specific Inference Heads:} The embedding $z$ is shared across four lightweight, independent prediction heads, ensuring that the inference pipeline remains low-latency ($\approx$ 10ms) for real-time control.

\begin{itemize}
    \item \textbf{Contact State ($c, s$):} We treat contact detection and slip detection as binary classification problems. These are modeled by logistic regression heads $\sigma(W_c^T z + b_c)$ and $\sigma(W_s^T z + b_s)$, trained via binary cross-entropy loss ($\mathcal{L}_{BCE}$).
    \item \textbf{Continuous States ($x, F$):} Contact location and normal force are regressed using small Multi-Layer Perceptrons (MLP) consisting of three hidden layers with ReLU activations. These are trained to minimize the Mean Squared Error (MSE), $\mathcal{L}_{MSE} = ||y_{pred} - y_{gt}||^2$.
\end{itemize}

\vspace{2mm}
\noindent \textbf{3) Joint Optimization:} The system is trained in a multi-task fashion on a labeled dataset $\mathcal{D} = \{(A_i, c_i, s_i, x_i, F_i)\}$. The total loss is a weighted sum of the individual task losses:
\begin{equation*}
\mathcal{L}_{total} = \lambda_c \mathcal{L}_{contact} + \mathbb{1}_{c=1} [\lambda_s \mathcal{L}_{slip} + \lambda_x \mathcal{L}_{loc} + \lambda_F \mathcal{L}_{force}]
\end{equation*}
We apply a masking term $\mathbb{1}_{c=1}$ to ensure that regression gradients are only backpropagated when contact is present, stabilizing the training dynamics for the continuous variables. The model is trained on real-world measurements augmented with additive noise to improve robustness. 




%% file: text/4_system_design.tex
\section{System design}

\subsection{Sensor hardware design}
Our active acoustic tactile sensor is inspired by string instruments such as guitars and violins, where contact location and force are reflected in the resulting acoustic response. We leverage both this physical intuition and commercially available string hardware to construct a compact and robust sensing system.

\begin{figure}
\centering
\includegraphics[width=0.95\linewidth]{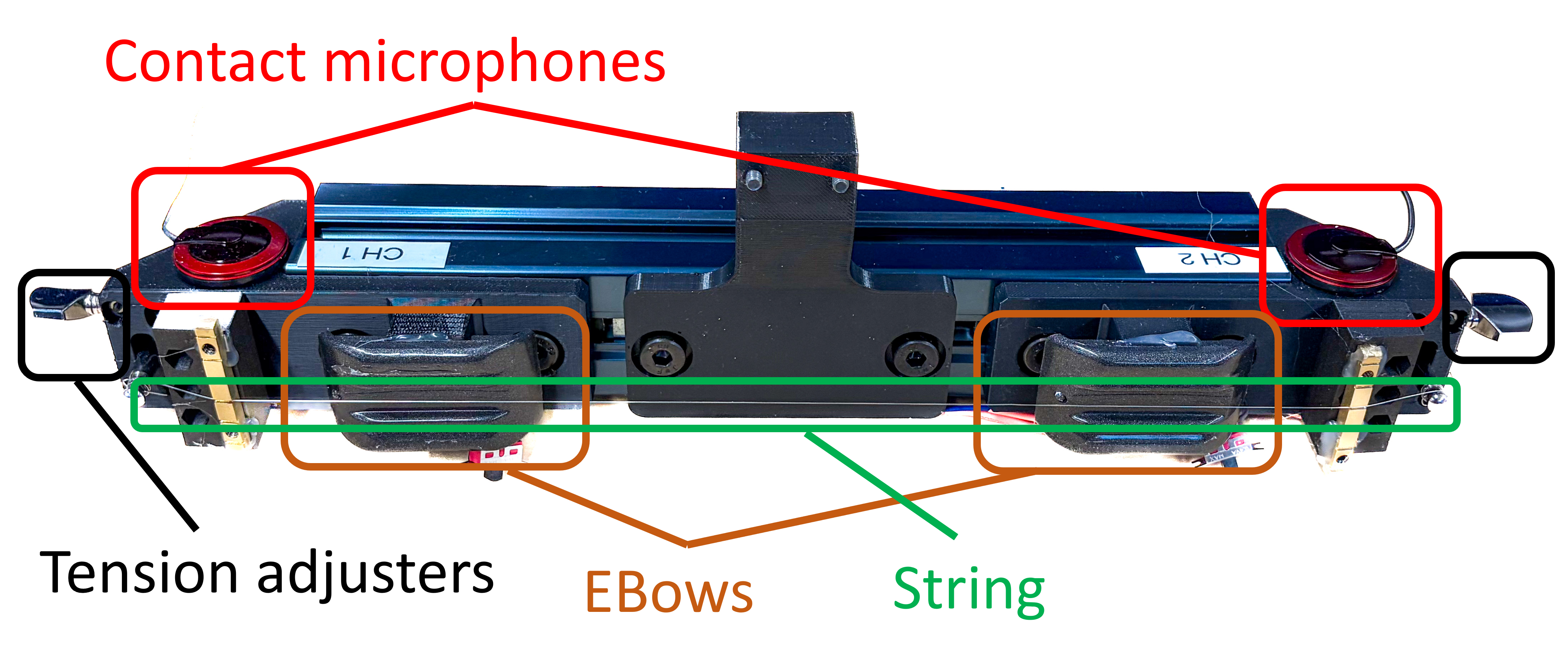}
\caption{\textbf{Hardware design of the string-based tactile sensor.} The system integrates an aluminum frame supporting a steel guitar string, dual electromagnetic EBows for continuous excitation, and contact microphones for high-frequency vibration acquisition.}
\label{fig:violin-prototypes}
\end{figure}

As shown in Fig. \ref{fig:violin-prototypes}, the sensor consists of five components: a rigid frame, tension adjusters, a tensioned string, a continuous excitation mechanism, and contact microphones. The main structure is built from an aluminum bar to ensure stable string tension over time. Custom 3D-printed fixtures are used for string support, tension adjustment, and mounting the excitation hardware. 
String and Tension adjusters are off-the-shelf guitar components, and are reliable and easy to replace.

We use a commercially available steel guitar string as the vibrating element. Continuous excitation is provided by an EBow~\cite{heet1978ebow}, an electromagnetic driver that sustains string vibration via feedback between sensing and excitation coils and preferentially excites resonant modes. The EBow is mechanically modified for integration into the sensor. Vibration signals are captured using contact microphones adapted from guitar hardware, enabling robust acquisition of high-frequency contact-induced responses. The sensing range is 10 cm.

\subsection{System setup}
The system setup is designed for an end-to-end signal chain from the sensor to real-time inference on a desktop computer. Here we describe the audio acquisition and processing stack that connects the two contact microphones of the sensor to the host machine for low-latency streaming and inference.

\begin{figure}
    \centering
    \includegraphics[width=0.98\linewidth]{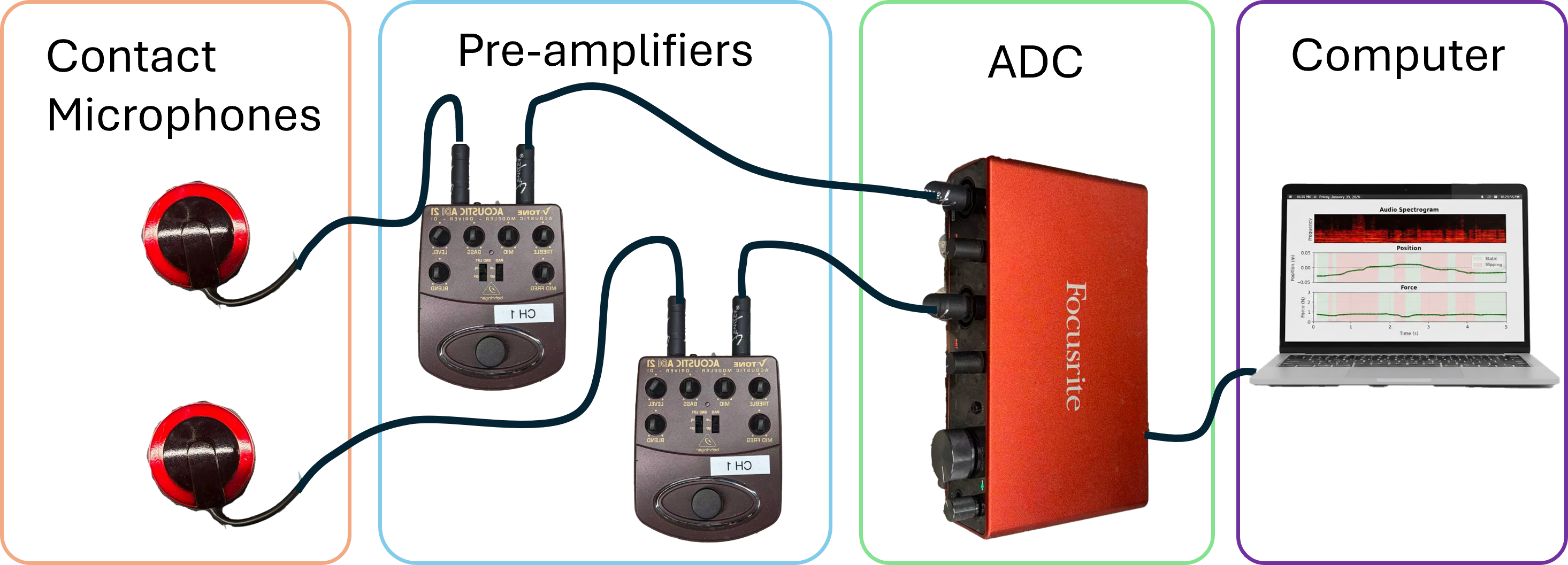}
    \caption{System-level audio acquisition pipeline. Audio signals from two contact microphones are independently amplified, digitized by a multi-channel audio interface, and streamed to a desktop computer for real-time processing.}
  \label{fig:system_setup}
  \vspace{-0.4cm}
\end{figure}

Each microphone is routed through an independent acoustic preamplifier (ADL21 Acoustic Amp Modeler/Direct Recording preamp) to provide impedance matching and gain prior to digitization, Fig.~\ref{fig:system_setup}. The two amplified signals are then recorded as separate inputs on a Focusrite Scarlett 4i4 audio interface (ADC) and streamed to the computer over USB. 

On the host machine, audio is captured using a low-latency JACK audio setup \cite{letz2005jack} to enable stable real-time streaming and synchronized two-channel recording. All experiments are conducted at a sampling rate of 44.1~kHz, with predictions computed from short sliding audio windows to enable real-time operation. 

\subsection{Design choices and parameter sensitivity}
We analyze several key design choices to better understand the trade-offs between accuracy, robustness, and latency, and to demonstrate that the proposed sensing modality is not sensitive to specific modeling or learning decisions.
\subsubsection{Feature Extraction}
We evaluate two complementary feature representations: AST-based features and FFT harmonic Features. \textbf{AST-based features:} we use a pretrained Audio Spectrogram Transformer (AST) to extract high-level representations from each audio channel. Raw audio is resampled to 16 kHz, converted to Mel spectrograms, and passed through the transformer encoder. Mean and max pooling over time yield a 1536-D embedding per channel, concatenated into a 3072-D feature vector per audio piece. AST features provide robustness to noise and capture global spectral structure.

\textbf{FFT harmonic features:} For low-latency operation, we also extract explicit harmonic features using short-time Fourier transforms. For FFT-based features, we extract band energy around the first 16 harmonics from each microphone channel, as well as cross-channel difference and ratio features. These four groups, Mic 1, Mic 2, inter-channel differences, and inter-channel ratios, are concatenated to form a 300D feature vector. These 300D features consists of These features directly encode physically interpretable vibration cues.

\subsubsection{Data collection and Augmentation}
Using the system setup we described before, we collected 10,791 labeled static contact audio samples, and 6,795 labeled slipping contact audio samples, each corresponding to a distinct contact position and force. Data are temporally split into training and testing sets to evaluate generalization across sessions. 

To improve robustness, we apply noise-only augmentation during training.
We augment each window by injecting randomly sampled noise at 5–25 dB SNR, including white Gaussian noise, pink (1/f) noise, 60 Hz electrical hum, and band-limited high-frequency noise (2–8 kHz).
Importantly, we avoid pitch or time scaling to preserve spatial frequency cues critical for localization.

\subsubsection{Window length of audio}
The choice of window length involves a trade-off between spectral resolution and system latency. While longer windows improve accuracy by providing more stable harmonic features, they increase the sensing lag; conversely, shorter windows reduce latency but may compromise frequency resolution. We selected a 0.1\,[sec] window as an optimal balance. This duration is sufficient to capture the resonant modes required for millimeter-scale localization and reliable force estimation while maintaining the low-latency response necessary for real-time slip detection and robotic interaction.

%% file: text/5_experiment.tex
\section{Experiments}
\subsection{Experiment setup and procedure}

\begin{figure}[h!]
    \centering
    \includegraphics[width=0.7\linewidth]{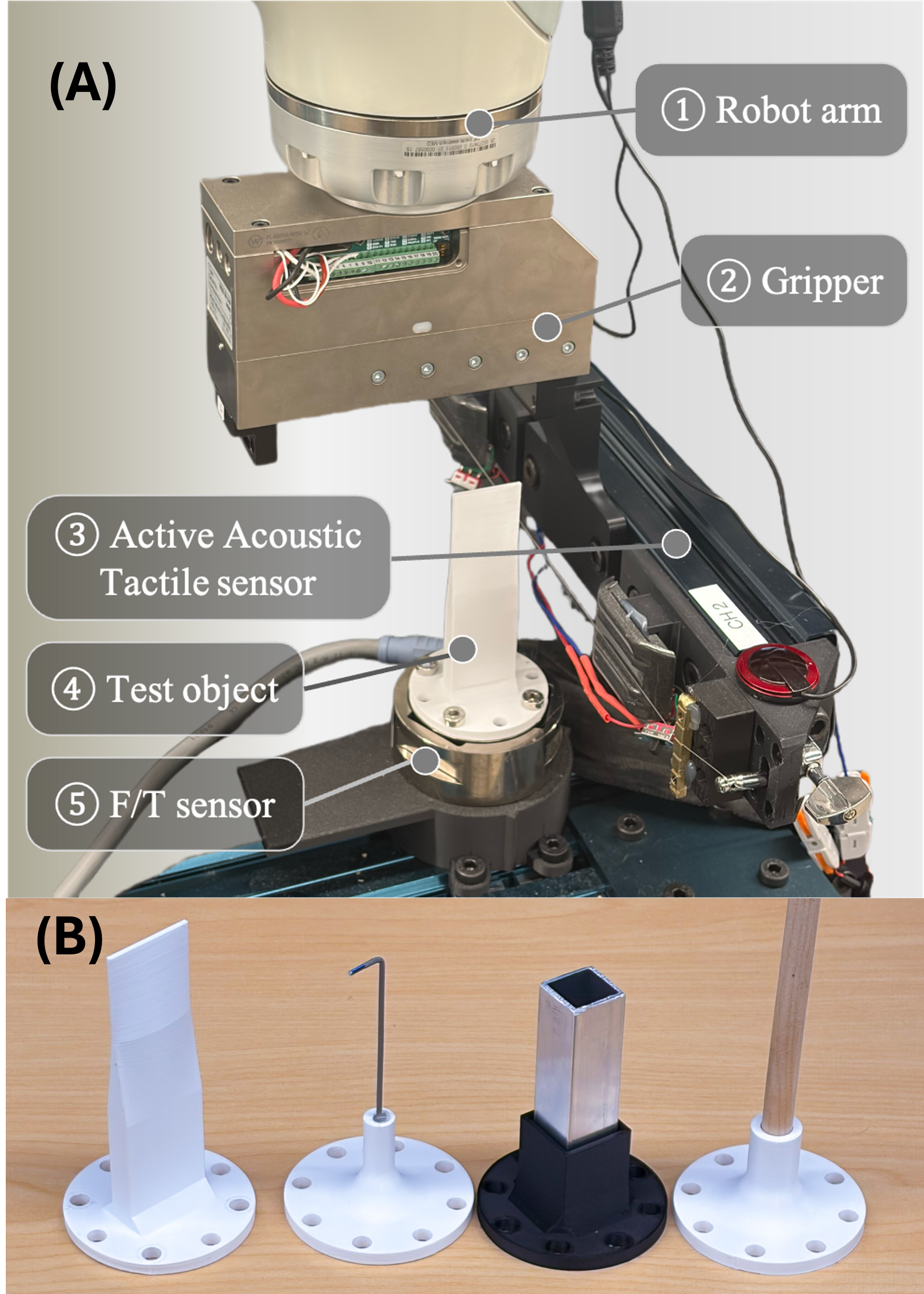}
    \caption{\textbf{Experiment setup.}  (A) The sensor is mounted on a KUKA LBR Med R820 robot with a WSG-50 gripper. A table-mounted ATI Gamma F/T sensor provides ground-truth force and geometry labels during data collection. (B) We collect training data using 3D printed object, and evaluate on allen key, metal tube, and wood sick.}
    \vspace{-3mm}
    \label{fig:experiment_setup}
\end{figure}

\textbf{Robot and gripper mounting.} We mount our tactile sensor on one finger of a WSG-50 parallel gripper, which is attached to a KUKA LBR Med R820 robot arm, as shown in Fig. \ref{fig:experiment_setup}. For data collection, the robot moves our sensor finger to make and break contact against a stationary object while maintaining a fixed approach direction. Audio is recorded and labelled with contact force, contact location, slippage/no slippage, and contact/no contact labels.

\textbf{Ground-truth force and contact geometry.} To obtain force measurements, an ATI Gamma force/torque sensor is rigidly mounted to the table. A 3D-printed fixture holding different contact profiles is bolted to the ATI sensor, enabling repeatable contacts with different surface radius. During data collection, we use a single 3D-printed object that contains multiple regions with different radii; the robot contacts different regions to sweep the radius condition while keeping the fixture and coordinate frame fixed. For evaluation, we have 3 unseen objects: metal tube, wood stick, and an allen key.

\textbf{Labels.} Each sample consists of the synchronized dual-channel audio window $A$, the applied normal force $F$ from the ATI sensor, and the contact location $x$ along the string. We obtain $x$ from the robot's proprioception and calibrated force/torque sensor frame. No-contact windows are collected by repeating the same motion trajectories with a clearance offset. For slippage dataset, we collect audio signal while moving with string contacts the object, ranging from 0 to 0.03 m/s, in both horizontal and vertical directions.

\textbf{Parameter sweep.} We collect data by scanning over a grid of conditions spanning a range of contact depth, contact locations along the string, and surface radii. This produces paired datasets that cover both steady contacts and transitions between contact/no-contact, which are later used for contact detection, regression, and slip analysis.
\subsection{Evaluation}

Performance is assessed using a held-out test set consisting of the original 3D-printed calibration object and three additional unseen rigid objects: a metal tube, a wooden stick, and an Allen key. Contact location and normal force estimation are evaluated using Mean Absolute Error (MAE), task-relevant accuracy thresholds, and the Pearson correlation coefficient. To ensure robustness across diverse operating conditions, all results are computed on test trajectories that include both contact and no-contact segments, and we additionally evaluate the system's performance under external acoustic disturbances.

\subsection{Environmental noise robustness}
To evaluate robustness against external acoustic interference, we introduced a dynamic noise source (pop music) via a speaker positioned 20 cm from the sensor at approximately 50\% volume. We collected 70 s of labeled data using the same ground-truth setup (ATI sensor and robot kinematics) and compared the system's performance on this noisy dataset against the noise-free baseline.

\subsection{Data analysis}
\begin{figure}[h!]
    \centering
    \includegraphics[width=1\linewidth]{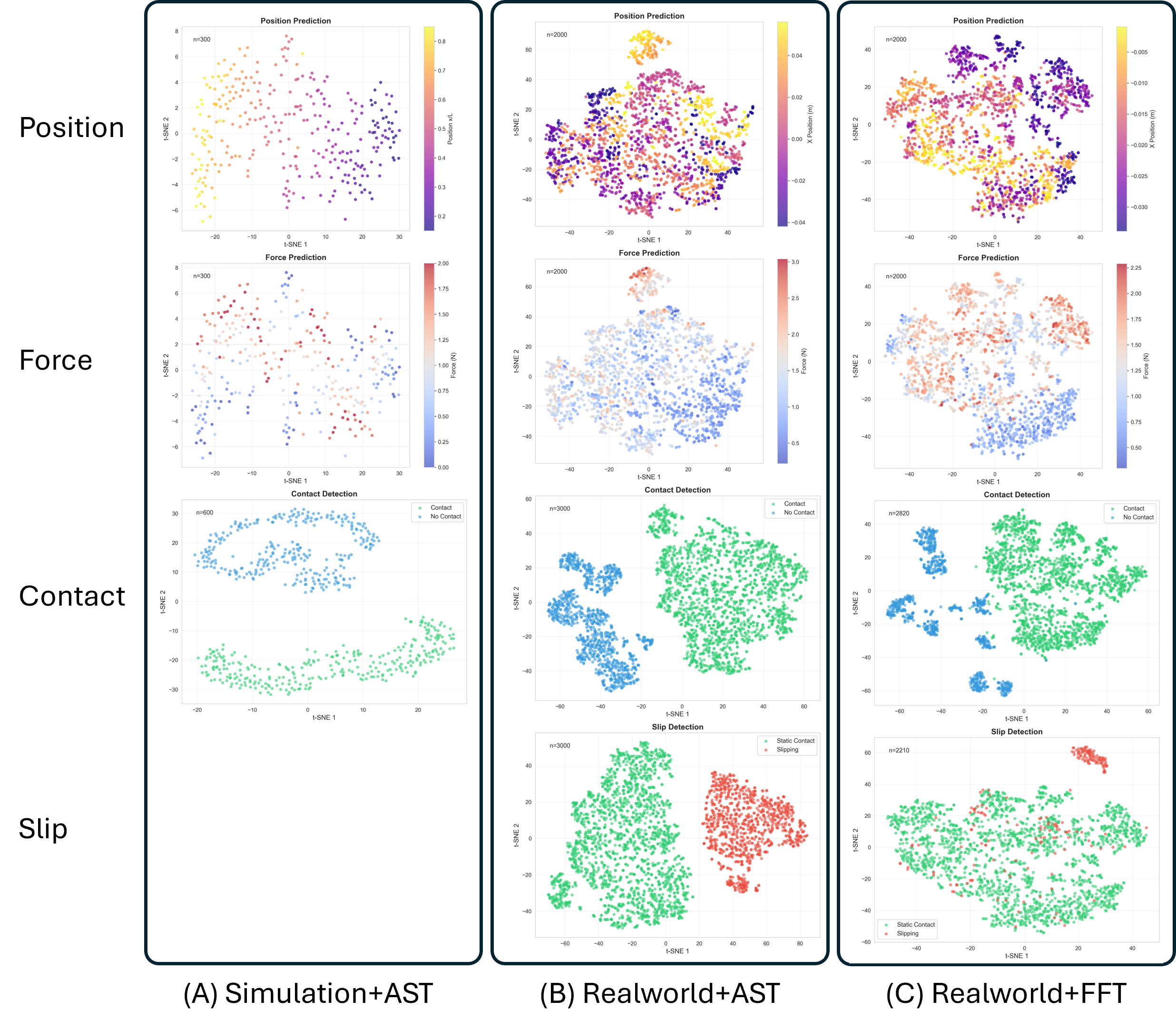}
    \caption{\textbf{t-SNE visualization of audio feature spaces.} 
    We compare (A) Simulated AST features, (B) Real-world AST features, and (C) Real-world FFT features across four tactile tasks. 
    Simulated data (A) shows a highly organized structure, indicating that tactile relationships are easily learnable in ideal conditions. 
    Real-world embeddings (B, C) are more complex due to physical phenomena like overtones and damping. 
    AST features (B) significantly outperform FFT (C) in slip detection because the transformer architecture captures critical temporal dynamics that static spectral descriptors miss. 
    Contact detection remains robustly separated across all modalities.}
    \vspace{-0.3cm}
    \label{fig:data}
\end{figure}

We qualitatively analyze the structure of our dataset by visualizing feature embeddings using t-SNE. We compare (i) embeddings from an Audio Spectrogram Transformer (AST) and (ii) compact FFT magnitude features (300D). Each audio segment is mapped to a fixed-length feature vector and projected to 2D for inspection. While t-SNE does not preserve global distances, it is useful for diagnosing whether a representation yields separable clusters for classification or smooth organization for regression.

Fig.~\ref{fig:data} highlights clear representation-dependent trends. For \textbf{contact detection}, both AST and FFT features form well-separated groups, suggesting that contact introduces a strong acoustic signature and that the classification boundary is simple in either feature space. In contrast, \textbf{slip detection} shows substantial overlap between static and slipping samples, with FFT features exhibiting particularly strong mixing. This suggests that slip-related signatures are comparatively subtle and may depend on \emph{spectro-temporal} patterns (e.g., transient bursts or evolving harmonics) that are not well represented by a single static spectral-magnitude descriptor, whereas AST features can better preserve such structure.

For the continuous targets, we observe that AST features induce more structured organization than FFT. \textbf{Force} exhibits a more coherent ordering in the embedding, whereas \textbf{contact location} is noticeably more mixed and does not display a strong global monotonic trend, consistent with the increased ambiguity of inferring position from audio alone.

Overall, these visualizations motivate our modeling choice: rather than relying on heavy end-to-end architectures, we use lightweight prediction heads on top of informative representations. The strong separability for contact and the structured organization for force (particularly with AST) suggest that simple classifiers/regressors are sufficient to achieve high performance, while slip and fine-grained location remain the more challenging regimes.

%% file: text/6_results.tex
\section{Experiment Results}

We evaluate binary classification for contact occurrence and slippage detection using lightweight classifiers on top of extracted audio features. On the evaluated test set, both tasks achieve perfect classification accuracy. Rather than reporting tabular metrics, we analyze the structure of the feature space to better understand task difficulty.

As shown in Fig. \ref{fig:data}, static contact and slipping contact form clearly separated clusters in the learned feature space, with a large margin and no observable overlap. This indicates that slipping contact produces a strong and distinctive acoustic signature compared to static contact, making the classification problem effectively linearly separable. As a result, simple classifiers are sufficient to achieve perfect performance without relying on complex models or task-specific tuning. 
It is important to note that slip detection was evaluated under controlled, low-velocity conditions (constant speed $v \le 0.03$ m/s). While the current model achieves robust detection in this regime, highly dynamic slip events characterized by rapid acceleration or higher velocities may exhibit distinct spectral transient properties. Future work will investigate the system's generalization to these high-speed dynamic interactions.

Table \ref{tab:location_noise} and Table \ref{tab:force_noise} summarize contact location and force estimation performance across objects under clean and noisy conditions. On rigid objects such as plastic, metal tube, and the Allen key, the proposed sensor achieves millimeter-scale contact localization, with mean absolute error on the order of 2–5 mm and over 70–88\% of predictions within 5 mm. Pearson correlation exceeds 0.95 in these cases, indicating a strong and consistent relationship between string vibration features and contact position. Localization performance degrades on wood, where error increases to approximately 9 mm and accuracy within 5 mm drops to around 50\%. This behavior is expected, as wood is less rigid and more acoustically damped, attenuating high-frequency string vibrations and reducing the signal-to-noise ratio of contact-induced spectral features.

Force estimation is inherently more challenging than position estimation, yet the system achieves mean absolute errors of 0.11–0.17 N, with 67–84\% of predictions within 0.2 N across objects (Max force 2N). One contributing factor is the physical coupling between the vibrating string and the force–torque sensor: during contact, the string continues to oscillate and transmits small dynamic force components to the object, introducing variance in the measured ground-truth force. This variance exists in both training and test data and manifests as a systematic noise floor rather than random error. Despite these challenges, both location and force estimation remain robust to external acoustic noise, with only minor degradation in accuracy and correlation, demonstrating that the learned representations primarily capture intrinsic vibration patterns rather than ambient sound.

\input{table/results_tab}


%% file: table/results_tab.tex
\begin{table}[t]
\centering
\caption{Contact location estimation robustness under external acoustic noise.
Performance is reported using mean absolute error (MAE), percentage of predictions
within 5\,mm, and Pearson correlation coefficient for clean and noisy conditions.
All results use AST features with a 0.1\,s audio window. }
\label{tab:location_noise}
\begin{tabular}{lcccc}
\toprule
\textbf{Object} &
\textbf{MAE (mm)} &
\textbf{$\le$5 mm (\%)} &
\textbf{Pearson $r$} &
\textbf{Condition} \\
\midrule
Plastic   & 2.7 & 87.8 & 0.994 & Clean \\
          & 3.1 & 83.3 & 0.994 & Noise \\
\midrule
Wood      & 8.8 & 51.4 & 0.863 & Clean \\
          & 8.6 & 55.7 & 0.857 & Noise \\
\midrule
Metal tube  & 5.4 & 70.0 & 0.952 & Clean \\
          & 8.0 & 57.1 & 0.893 & Noise \\
\midrule
Allen Key & 2.7 & 82.9 & 0.993 & Clean \\
          & 2.8 & 87.1 & 0.992 & Noise \\
\bottomrule
\end{tabular}
\end{table}

\vspace{-3mm}

\begin{table}[t]
\centering
\caption{Contact force estimation robustness under external acoustic noise.
Performance is evaluated using mean absolute error (MAE), percentage of predictions
within 0.2\,N, and Pearson correlation coefficient for clean and noisy conditions.
All results use AST features with a 0.1\,s audio window. }
\label{tab:force_noise}
\begin{tabular}{lcccc}
\toprule
\textbf{Object} &
\textbf{MAE (N)} &
\textbf{$\le$0.2 N (\%)} &
\textbf{Pearson $r$} &
\textbf{Condition} \\
\midrule
Plastic   & 0.141 & 74.4 & 0.909 & Clean \\
          & 0.131 & 75.6 & 0.928 & Noise \\
\midrule
Wood      & 0.111 & 84.3 & 0.903 & Clean \\
          & 0.115 & 82.9 & 0.927 & Noise \\
\midrule
Metal tube  & 0.172 & 70.0 & 0.908 & Clean \\
          & 0.173 & 65.7 & 0.906 & Noise \\
\midrule
Allen Key & 0.167 & 67.1 & 0.844 & Clean \\
          & 0.165 & 65.7 & 0.819 & Noise \\
\bottomrule
\end{tabular}
\vspace{-5mm}
\end{table}

%% file: text/7_limitation.tex
\section{Discussion and Limitations}


This work explores a continuous tactile sensing modality using tensioned strings and audio measurements. While the proposed approach demonstrates promising performance, it also introduces several limitations that suggest directions for future work.
First, sensing is inherently one-dimensional along the string, and the current system assumes a single dominant contact; extending to multiple contacts or higher-dimensional coverage would require additional mechanical and signal complexity. Second, string tension drift and hardware geometry influence frequency responses, necessitating calibration and limiting long-term stability. Third, highly compliant or heavily damped objects can attenuate vibrations, reducing signal quality. Finally, the current results focus on static or quasi-static contacts, as the string requires a short settling time to reach steady vibration, which poses challenges for highly dynamic interactions, especially for contact position and force sensing, though slip detection is instantaneous.

Despite these limitations, the approach offers compelling advantages. The hardware is  lightweight, scalable, can capture high-frequency signal, and only relying on simple components such as strings, electromagnetic drivers, and contact microphones. The sensing modality is robust to visual occlusion and surface contamination, supports real-time inference once steady vibration is established, and enables reliable contact localization, force estimation, and slip detection using high-frequency vibration cues. These properties make wave-based tactile sensing a promising complement to conventional tactile sensors for manipulation tasks.

%% file: Appendix.tex
\section*{Appendix}
\addcontentsline{toc}{section}{Appendix}

\section{Audio Simulation: Convergence Example}
\label{app:sim_convergence}

\begin{figure}[h!]
    \centering
    \includegraphics[width=\linewidth]{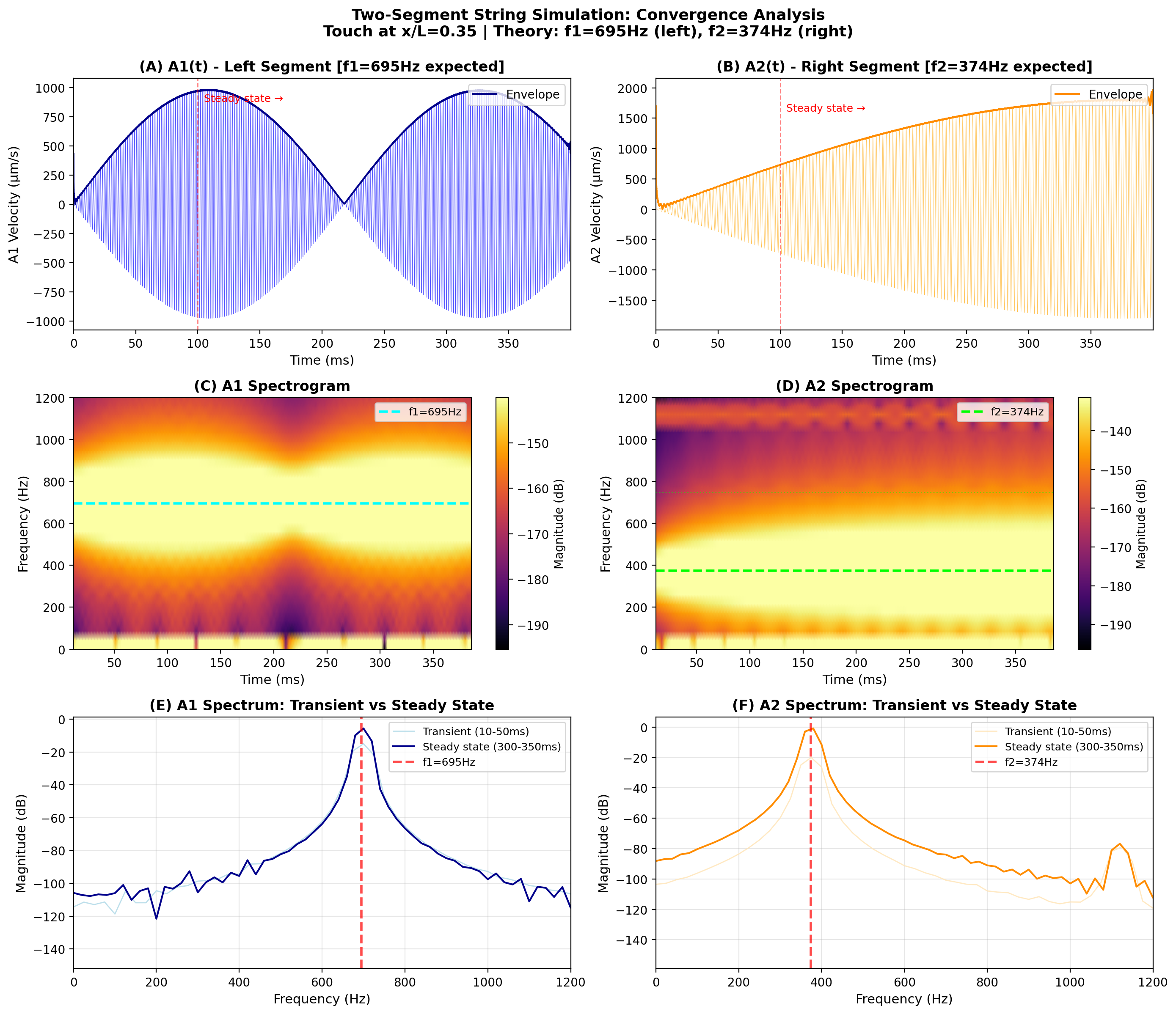}
    \caption{\textbf{Simulation convergence example.}
Two-segment string simulation for a contact at $x/L = 0.35$.
(A,B) Time-domain velocity signals with envelopes indicating convergence to steady oscillation.
(C,D) Spectrograms showing stable dominant frequency components.
(E,F) FFT spectra computed over transient (10--50~ms) and steady-state (300--350~ms) windows.}
    \label{fig:sim_convergence}
\end{figure}

This section provides additional context for Fig.~\ref{fig:sim_convergence}, which illustrates the convergence behavior of the string-vibration simulator under a representative contact condition. The purpose of this figure is to verify that the numerical integration and excitation scheme converge to stable, interpretable vibration modes before spectral features are extracted.

Fig.~\ref{fig:sim_convergence}(A,B) show the simulated transverse velocity signals for the two effective string segments following a contact at $x/L = 0.35$. The overlaid envelopes indicate a brief transient phase followed by convergence to steady oscillations. Panels (C,D) visualize the same signals in the time--frequency domain, confirming that dominant frequency components remain stable after convergence. Panels (E,F) compare FFT spectra computed over a transient window (10--50~ms) and a steady-state window (300--350~ms), demonstrating spectral sharpening and stabilization once steady oscillation is reached
.

This convergence analysis motivates discarding an initial transient period and using short steady-state windows for feature extraction in both simulation analysis and real-world audio processing.

\subsection{Simulation Parameters for Fig.~\ref{fig:sim_convergence}}

The parameters used to generate Fig.~\ref{fig:sim_convergence} are listed below. Values were selected to produce clear modal separation and are not tuned to match any specific hardware instance.

\begin{itemize}
    \item Total string length: $L = 0.65$~m
    \item Contact location: $x/L = 0.35$
    \item Linear density: $\mu = 6.5 \times 10^{-4}$~kg/m
    \item Baseline tension: $T_0 = 65$~N
    \item Force--tension coupling: $k_T = 6$~N/N
    \item Damping coefficient (no contact): $d_0 = 0.02$
    \item Spatial discretization: $N = 256$ nodes per segment
    \item Time step: $\Delta t = 1.0~\mu$s
    \item Simulation duration: 400~ms
\end{itemize}

Theoretical fundamental frequencies predicted by Eq.~(2) for this configuration are $f_1 = 695$~Hz and $f_2 = 374$~Hz, shown as dashed lines in the spectrogram and FFT panels. Minor deviations and higher-order components arise from damping and finite discretization effects.

\section{Real-world audio details}
\label{app:real_audio}

A key reason we rely on data-driven inference rather than purely analytical, peak-tracking methods is that the real string--driver system is \emph{multi-stable}: even under the same hardware configuration (string tension, EBow placement, pickup locations) and the same nominal ``no-contact'' condition, the steady-state vibration can settle into qualitatively different spectral patterns. In practice, the EBow--string interaction can lock into different combinations of harmonics and sidebands depending on the transient history of the system. As a result, the observed spectrum is not a single deterministic function of the current contact state alone, but also depends on the recent excitation and decay trajectory.

\begin{figure}[h!]
    \centering
    \includegraphics[width=0.95\linewidth]{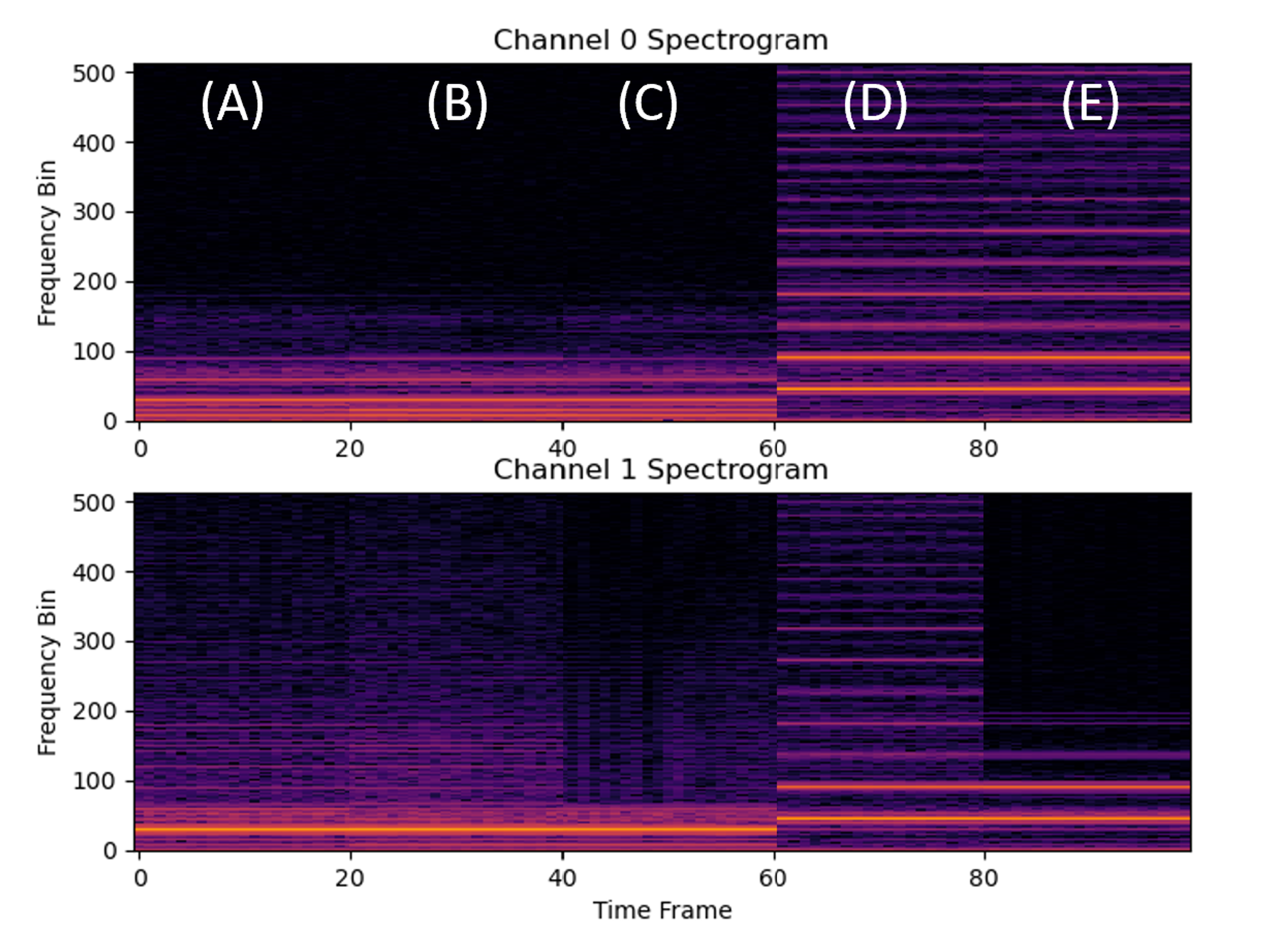}
    \caption{\textbf{Multiple steady-state vibration modes in real audio (no contact).}
    Two-channel spectrograms recorded from the same hardware in nominally identical no-contact conditions. (A)--(E) show five distinct steady-state spectral patterns that can arise due to different excitation histories (e.g., after contacts at different locations/forces followed by disengagement). This multi-stability motivates robust, data-driven feature extraction rather than deterministic peak-tracking.}
    \vspace{-3mm}
    \label{fig:multiple_modes}
\end{figure}

Fig.~\ref{fig:multiple_modes} illustrates this phenomenon. Across five trials with no contact, the two-channel spectrograms exhibit distinct harmonic structures and relative energy distributions, despite identical sensing hardware. These modes are commonly triggered by different initial conditions, e.g., after an object makes contact at different locations/forces and then disengages. This mode variability makes it difficult to define a robust hand-engineered mapping from a small set of spectral peaks to contact quantities. Instead, we use learned representations that are trained to be tolerant to such variability while still preserving the contact-dependent cues needed for localization, force estimation, and slip detection.

%% file: references.bib
@article{yuan2017gelsight,
  title={Gelsight: High-resolution robot tactile sensors for estimating geometry and force},
  author={Yuan, Wenzhen and Dong, Siyuan and Adelson, Edward H},
  journal={Sensors},
  volume={17},
  number={12},
  pages={2762},
  year={2017},
  publisher={Multidisciplinary Digital Publishing Institute}
}

@article{li2025classification,
  title={Classification of Vision-Based Tactile Sensors: A Review},
  author={Li, Haoran and Lin, Yijiong and Lu, Chenghua and Yang, Max and Psomopoulou, Efi and Lepora, Nathan F},
  journal={IEEE Sensors Journal},
  year={2025},
  publisher={IEEE}
}

@article{cirillo2021tactile,
  title={Tactile sensors for parallel grippers: Design and characterization},
  author={Cirillo, Andrea and Costanzo, Marco and Laudante, Gianluca and Pirozzi, Salvatore},
  journal={Sensors},
  volume={21},
  number={5},
  pages={1915},
  year={2021},
  publisher={MDPI}
}

@article{pattabiraman2025eflesh,
  title={eFlesh: Highly customizable Magnetic Touch Sensing using Cut-Cell Microstructures},
  author={Pattabiraman, Venkatesh and Huang, Zizhou and Panozzo, Daniele and Zorin, Denis and Pinto, Lerrel and Bhirangi, Raunaq},
  journal={arXiv preprint arXiv:2506.09994},
  year={2025}
}

@inproceedings{bhirangi2025anyskin,
  title={Anyskin: Plug-and-play skin sensing for robotic touch},
  author={Bhirangi, Raunaq and Pattabiraman, Venkatesh and Erciyes, Enes and Cao, Yifeng and Hellebrekers, Tess and Pinto, Lerrel},
  booktitle={2025 IEEE International Conference on Robotics and Automation (ICRA)},
  pages={16563--16570},
  year={2025},
  organization={IEEE}
}

@article{meribout2024tactile,
  title={Tactile sensors: A review},
  author={Meribout, Mahmoud and Takele, Natnael Abule and Derege, Olyad and Rifiki, Nidal and El Khalil, Mohamed and Tiwari, Varun and Zhong, Jing},
  journal={Measurement},
  volume={238},
  pages={115332},
  year={2024},
  publisher={Elsevier}
}

@article{suhn2023vibro,
  title={Vibro-Acoustic sensing of instrument interactions as a potential source of texture-related information in robotic palpation},
  author={S{\"u}hn, Thomas and Esmaeili, Nazila and Mattepu, Sandeep Y and Spiller, Moritz and Boese, Axel and Urrutia, Robin and Poblete, Victor and Hansen, Christian and Lohmann, Christoph H and Illanes, Alfredo and others},
  journal={Sensors},
  volume={23},
  number={6},
  pages={3141},
  year={2023},
  publisher={MDPI}
}

@inproceedings{donlon2018gelslim,
  title={Gelslim: A high-resolution, compact, robust, and calibrated tactile-sensing finger},
  author={Donlon, Elliott and Dong, Siyuan and Liu, Melody and Li, Jianhua and Adelson, Edward and Rodriguez, Alberto},
  booktitle={2018 IEEE/RSJ International Conference on Intelligent Robots and Systems (IROS)},
  pages={1927--1934},
  year={2018},
  organization={IEEE}
}

@article{taylor2021gelslim3,
  title={GelSlim3. 0: High-Resolution Measurement of Shape, Force and Slip in a Compact Tactile-Sensing Finger},
  author={Taylor, Ian and Dong, Siyuan and Rodriguez, Alberto},
  journal={arXiv preprint arXiv:2103.12269},
  year={2021}
}

@article{kuppuswamy2020soft,
  title={Soft-Bubble grippers for robust and perceptive manipulation},
  author={Kuppuswamy, Naveen and Alspach, Alex and Uttamchandani, Avinash and Creasey, Sam and Ikeda, Takuya and Tedrake, Russ},
  journal={arXiv preprint arXiv:2004.03691},
  year={2020}
}

@patent{heet1978ebow,
  title     = {Hand-held electronic device for producing sustained musical tones},
  author    = {Heet, Gregory J.},
  year      = {1978},
  month     = feb,
  number    = {US4075921A},
  type      = {U.S. Patent},
  url       = {https://patents.google.com/patent/US4075921A}
}

@inproceedings{lu2023active,
  title={Active acoustic sensing for robot manipulation},
  author={Lu, Shihan and Culbertson, Heather},
  booktitle={2023 IEEE/RSJ International Conference on Intelligent Robots and Systems (IROS)},
  pages={3161--3168},
  year={2023},
  organization={IEEE}
}

@inproceedings{jamali2010material,
  title={Material classification by tactile sensing using surface textures},
  author={Jamali, Nawid and Sammut, Claude},
  booktitle={2010 IEEE International Conference on Robotics and Automation},
  pages={2336--2341},
  year={2010},
  organization={IEEE}
}

@article{fernandez2014micro,
  title={Micro-vibration-based slip detection in tactile force sensors},
  author={Fernandez, Raul and Payo, Ismael and Vazquez, Andres S and Becedas, Jonathan},
  journal={Sensors},
  volume={14},
  number={1},
  pages={709--730},
  year={2014},
  publisher={Molecular Diversity Preservation International (MDPI)}
}

@inproceedings{taunyazov2021extended,
  title={Extended tactile perception: Vibration sensing through tools and grasped objects},
  author={Taunyazov, Tasbolat and Song, Luar Shui and Lim, Eugene and See, Hian Hian and Lee, David and Tee, Benjamin CK and Soh, Harold},
  booktitle={2021 IEEE/RSJ International Conference on Intelligent Robots and Systems (IROS)},
  pages={1755--1762},
  year={2021},
  organization={IEEE}
}

@inproceedings{letz2005jack,
  title={Jack audio server for multi-processor machines},
  author={Letz, St{\'e}phane and Orlarey, Yann and Fober, Dominique},
  booktitle={International Computer Music Conference},
  pages={1--4},
  year={2005}
}

@article{zhao2025universal,
  title={Universal slip detection of robotic hand with tactile sensing},
  author={Zhao, Chuangri and Yu, Yang and Ye, Zeqi and Tian, Ziyang and Zhang, Yifan and Zeng, Ling-Li},
  journal={Frontiers in Neurorobotics},
  volume={19},
  pages={1478758},
  year={2025},
  publisher={Frontiers Media SA}
}

@article{zhang2025vibecheck,
  title={VibeCheck: Using Active Acoustic Tactile Sensing for Contact-Rich Manipulation},
  author={Zhang, Kaidi and Kim, Do-Gon and Chang, Eric T and Liang, Hua-Hsuan and He, Zhanpeng and Lampo, Kathryn and Wu, Philippe and Kymissis, Ioannis and Ciocarlie, Matei},
  journal={arXiv preprint arXiv:2504.15535},
  year={2025}
}

@article{funk2024evetac,
  title={Evetac: An event-based optical tactile sensor for robotic manipulation},
  author={Funk, Niklas and Helmut, Erik and Chalvatzaki, Georgia and Calandra, Roberto and Peters, Jan},
  journal={IEEE Transactions on Robotics},
  year={2024},
  publisher={IEEE}
}

@inproceedings{yamaguchi2017implementing,
  title={Implementing tactile behaviors using fingervision},
  author={Yamaguchi, Akihiko and Atkeson, Christopher G},
  booktitle={2017 IEEE-RAS 17th International Conference on Humanoid Robotics (Humanoids)},
  pages={241--248},
  year={2017},
  organization={IEEE}
}

@inproceedings{howe1989sensing,
  title={Sensing skin acceleration for slip and texture perception.},
  author={Howe, Robert D and Cutkosky, Mark R},
  booktitle={ICRA},
  pages={145--150},
  year={1989}
}

@article{heyneman2016slip,
  title={Slip classification for dynamic tactile array sensors},
  author={Heyneman, Barrett and Cutkosky, Mark R},
  journal={The International Journal of Robotics Research},
  volume={35},
  number={4},
  pages={404--421},
  year={2016},
  publisher={SAGE Publications Sage UK: London, England}
}

@article{yamaguchi2019recent,
  title={Recent progress in tactile sensing and sensors for robotic manipulation: can we turn tactile sensing into vision?},
  author={Yamaguchi, Akihiko and Atkeson, Christopher G},
  journal={Advanced Robotics},
  volume={33},
  number={14},
  pages={661--673},
  year={2019},
  publisher={Taylor \& Francis}
}

@article{lambeta2020digit,
  title={Digit: A novel design for a low-cost compact high-resolution tactile sensor with application to in-hand manipulation},
  author={Lambeta, Mike and Chou, Po-Wei and Tian, Stephen and Yang, Brian and Maloon, Benjamin and Most, Victoria Rose and Stroud, Dave and Santos, Raymond and Byagowi, Ahmad and Kammerer, Gregg and others},
  journal={IEEE Robotics and Automation Letters},
  volume={5},
  number={3},
  pages={3838--3845},
  year={2020},
  publisher={IEEE}
}

@article{lambeta2024digitizing,
  title={Digitizing touch with an artificial multimodal fingertip},
  author={Lambeta, Mike and Wu, Tingfan and Sengul, Ali and Most, Victoria Rose and Black, Nolan and Sawyer, Kevin and Mercado, Romeo and Qi, Haozhi and Sohn, Alexander and Taylor, Byron and others},
  journal={arXiv preprint arXiv:2411.02479},
  year={2024}
}

@article{liu2024sonicsense,
  title={Sonicsense: Object perception from in-hand acoustic vibration},
  author={Liu, Jiaxun and Chen, Boyuan},
  journal={arXiv preprint arXiv:2406.17932},
  year={2024}
}

@article{sipos2024gelslim,
  title={GelSlim 4.0: Focusing on touch and reproducibility},
  author={Sipos, Andrea and Bogert, William van den and Fazeli, Nima},
  journal={arXiv preprint arXiv:2409.19770},
  year={2024}
}

@inproceedings{alspach2019soft,
  title={Soft-bubble: A highly compliant dense geometry tactile sensor for robot manipulation},
  author={Alspach, Alex and Hashimoto, Kunimatsu and Kuppuswamy, Naveen and Tedrake, Russ},
  booktitle={2019 2nd IEEE International Conference on Soft Robotics (RoboSoft)},
  pages={597--604},
  year={2019},
  organization={IEEE}
}

@inproceedings{padmanabha2020omnitact,
  title={Omnitact: A multi-directional high-resolution touch sensor},
  author={Padmanabha, Akhil and Ebert, Frederik and Tian, Stephen and Calandra, Roberto and Finn, Chelsea and Levine, Sergey},
  booktitle={2020 IEEE International Conference on Robotics and Automation (ICRA)},
  pages={618--624},
  year={2020},
  organization={IEEE}
}

@article{tomo2017covering,
  title={Covering a robot fingertip with uSkin: A soft electronic skin with distributed 3-axis force sensitive elements for robot hands},
  author={Tomo, Tito Pradhono and Schmitz, Alexander and Wong, Wai Keat and Kristanto, Harris and Somlor, Sophon and Hwang, Jinsun and Jamone, Lorenzo and Sugano, Shigeki},
  journal={IEEE Robotics and Automation Letters},
  volume={3},
  number={1},
  pages={124--131},
  year={2017},
  publisher={IEEE}
}

@inproceedings{tomo2016modular,
  title={A modular, distributed, soft, 3-axis sensor system for robot hands},
  author={Tomo, Tito Pradhono and Wong, Wai Keat and Schmitz, Alexander and Kristanto, Harris and Sarazin, Alexandre and Jamone, Lorenzo and Somlor, Sophon and Sugano, Shigeki},
  booktitle={2016 IEEE-RAS 16th International Conference on Humanoid Robots (Humanoids)},
  pages={454--460},
  year={2016},
  organization={IEEE}
}

@article{maaten2008visualizing,
  title={Visualizing data using t-SNE},
  author={Maaten, Laurens van der and Hinton, Geoffrey},
  journal={Journal of machine learning research},
  volume={9},
  number={Nov},
  pages={2579--2605},
  year={2008}
}
